\documentclass[10pt,twocolumn]{article}

\usepackage[a4paper,margin=1.8cm]{geometry}
\usepackage{amsmath,amssymb,amsfonts}
\usepackage{graphicx}
\usepackage{url}

\title{Bridging Theory and Practice in Crafting Robust Spiking Reservoirs}

\author{}

\date{}

\begin{document}

\twocolumn[
\maketitle
\begin{center}
\begin{tabular}{c c}
\textbf{Ruggero Freddi} & \textbf{Nicolas Seseri} \\
\textit{Research and Development Department} & \textit{Research and Development Department} \\
\textit{Manava.plus} & \textit{Manava.plus} \\
Milan, Italy & Milan, Italy \\
\texttt{ruggero.freddi@manava.plus} & \texttt{nicolas.seseri@manava.plus} \\

\textbf{Diana Nigrisoli} & \textbf{Alessio Basti} \\
\textit{Research and Development Department} & \textit{Department of Engineering and Geology} \\
\textit{Manava.plus} & \textit{“G'. d'Annunzio” University of Chieti-Pescara} \\
Milan, Italy & Pescara, Italy \\
\texttt{diana.nigrisoli@manava.plus} & \texttt{alessio.basti@unich.it} \\
\end{tabular}
\end{center}
\vspace{1em}
]

\begin{abstract}
Spiking reservoir computing provides an energy-efficient approach to temporal processing, but reliably tuning reservoirs to operate at the edge-of-chaos is challenging due to experimental uncertainty. This work bridges abstract notions of criticality and practical stability by introducing and exploiting the robustness interval, an operational measure of the hyperparameter range over which a reservoir maintains performance above task-dependent thresholds. Through systematic evaluations of Leaky Integrate-and-Fire (LIF) architectures on both static (MNIST) and temporal (synthetic Ball Trajectories) tasks, we identify consistent monotonic trends in the robustness interval across a broad spectrum of network configurations: the robustness-interval width decreases with presynaptic connection density $\beta$ (i.e., directly with sparsity) and directly with the firing threshold $\theta$. We further identify specific $(\beta, \theta)$ pairs that preserve the analytical mean-field critical point $w_{\text{crit}}$, revealing iso-performance manifolds in the hyperparameter space. Control experiments on Erdős–Rényi graphs show the phenomena persist beyond small-world topologies. Finally, our results show  that $w_{\text{crit}}$ consistently falls within empirical high-performance regions, validating $w_{\text{crit}}$ as a robust starting coordinate for parameter search and fine-tuning. To ensure reproducibility, the full Python code is publicly available.
\end{abstract}

\section{Introduction}

Reservoir computing with Spiking Neural Networks (SNNs), particularly Liquid State Machines (LSM), has recently gained increasing attention as a lightweight and biologically inspired framework for temporal signal processing, dynamical pattern recognition, and energy-efficient neuromorphic implementations \cite{zhang2023survey, alvarez2025optimizing}. 
Within this paradigm, the reservoir is a fixed recurrent network that projects inputs onto a high-dimensional state space, while only the readout layer (linear or mildly nonlinear) is trained \cite{maass2002liquid}. The quality of performance depends critically on the reservoir dynamics, which must be stable enough to avoid e.g. saturation or activity collapse, and sufficiently rich to ensure separability and memory \cite{alvarez2025separability, zhang2015digitallsm}.
A vast body of theoretical and empirical research suggests that optimal performance in reservoir computing, including spiking architectures, is achieved close to a critical regime, often referred to as the “edge of chaos” \cite{ivanov2021edge,woo2024characterization, bertschinger2004edge}. In this regime, the network is located near the transition between a quiescent state and a highly active regime, where small variations in the input can produce significant changes in collective behavior. While characterizing this regime is often analytically prohibitive, Leaky Integrate-and-Fire (LIF) networks represent a powerful, biologically inspired, and analytically tractable paradigm. In particular, mean-field theory yields mathematical expressions that link the average firing rate and critical synaptic weight to global hyperparameters such as connectivity, firing threshold, and external input \cite{brochini2016critical,levina2007soc}. While these results allow for the identification of the theoretical critical point, they leave relevant practical aspects of robust reservoir design open. Indeed, a useful reservoir does not need to operate exactly at the critical point, but rather within a \emph{region} of the hyperparameter space where performance remains high and stable despite modeling uncertainties, stochastic variability, and hardware constraints \cite{schrauwen2008robustness,hazan2010lsmrobust}. From this perspective, it is crucial not only to identify the location of the critical point, but also to quantify the extent of the hyperparameter region that sustains high performance, and to understand how this region changes as the structural properties of the reservoir vary. Although studies exist on the relationship between criticality and performance \cite{kinouchi2006optimal,woo2024characterization}, to the best of our knowledge there is no systematic analysis, over a broad range of conditions, that connects mean-field predictions of criticality to a quantitative measure of performance robustness with respect to the mean synaptic weight.

%
In this work, we address the gap between theoretical criticality and practical design by introducing an operational metric. We define a robustness interval over the mean synaptic weight $w$, representing the range in which different reservoir performance remain above functional thresholds. Our goal is to characterize the morphology of this high-performance region as hyperparameters vary, relating it to an analytical criticality reference \cite{freddi2025meanfield}. Unlike point-wise analytical estimates, this approach explicitly quantifies the safety margin available for reservoir tuning under uncertainty conditions.

In particular, we provide the following key contributions:
\begin{itemize}

\item Through different architectural configurations and two distinct (static/temporal) tasks (MNIST and synthetic Ball Trajectories), we uncover consistent monotonic trends: the width $\Delta$ of the robustness interval scales inversely with presynaptic connection density $\beta$, and directly with the firing threshold $\theta$ and, less prominently, with the current amplitude $I$.

\item We identify $(\beta, \theta)$ pairs that, by preserving the theoretical critical point, yield nearly overlapping normalized performance, revealing fundamental iso-performance manifolds in the hyperparameter space.

\item While our analyses leveraged small-world architectures, we also perform extensive control experiments using directed Erdős–Rényi graphs, suggesting that the observed phenomena are intrinsic to the dynamics rather than specific topologies.

\item To promote verifiability and reproducibility, we provide the full Python code at \url{https://github.com/RuggeroFreddi/LSM-Robustness.git}.
\end{itemize}

The remainder of this paper is as follows: Section 2 provides a comprehensive description of the network architecture, computational tasks, and the experimental plan, alongside our formal definitions of robustness and hyperparameter analysis. Section 3 reports the corresponding findings. Finally, Section 4 discusses the implications of our work.

\section{Methods}

\subsection{Network architecture} \label{first}
The networks used are LIF neuron networks with a small-world topology generated via the Watts--Strogatz algorithm \cite{watts1998collective}; a topology control using directed Erd\H{o}s--R\'enyi random graphs is also performed. The global hyperparameters of interest are the presynaptic connection density $\beta$, the firing threshold $\theta$, and the external input $I$. Each network contains $N$ neurons (MNIST: $N=1000$; Ball Trajectories: $N=2000$). The average degree is set to $2 \beta N$, with $\beta \in \{0.2,0.3,0.4\}$, while the rewiring probability is $0.2$. Connections are directed with a probability of $0.5$ in each direction. Synaptic weights are drawn from a Gaussian distribution with variable mean and a coefficient of variation of $20$ (MNIST) or $10$ (Ball Trajectories), with the lower value for the Ball Trajectories task to contain the reservoir's dynamic variability. The leak constant of each neuron is sampled from a log-normal distribution with mean $1/500$ and coefficient of variation $0.5$. The dynamics of the membrane potential $v_i(t)$ is given by
\begin{equation*}
\dot v_i(t) = \sum_n I\,\delta(t-t_{i,n}^{\text{ext}}) 
+ \sum_j \sum_n w_{ij}\,\delta(t-t_{j,n}^{\text{syn}})
- \alpha_i v_i(t),
\end{equation*}
where $t_{i,n}^{\text{ext}}$ is the time at which neuron $i$ receives its $n$-th external spike, and $t_{j,n}^{\text{syn}}$ is the time at which neuron $j$ emits its $n$-th spike. When $v_i(t)$ reaches the firing threshold value $\theta$, the neuron emits a spike. The firing thresholds for MNIST, $\theta \in \{2, 1.438, 1.157\}$, and for the Ball Trajectories task, $\theta \in \{2, 1.141, 1.117\}$, were set according to the $(\beta,\theta)$ equivalence in Sec.~\ref{pareq}, so as to obtain shifts of the operating regime comparable to those produced by the tested $\beta$ values. The external current is $I \in \{0.5,1.5,2\}$ and the refractory period is $T_{\mathrm{ref}} = 3$. In all the experiments, unless the corresponding hyperparameter is explicitly varied, we use the reference values $\beta_{\mathrm{ref}}=0.2$, $\theta_{\mathrm{ref}}=2$, and $I_{\mathrm{ref}}=2$. Between two consecutive examples, the potentials are reset based on two schemes: (i) a fixed value chosen randomly once; (ii) a value drawn randomly at each reset. In both cases, the reset is uniform in $[0,\theta]$. 

We note that, beyond the spiking threshold, the impact of $I$ on network dynamics saturates due to the reset mechanism.

\subsection{Computational Tasks and Encoding}
Two classification tasks are considered. 
\paragraph{MNIST}
The dataset is reduced to 6000 binary images, encoded via \emph{rate coding}: each pixel generates a spike train with a frequency proportional to its intensity. The task mainly requires input separability and does not require temporal memory.

\paragraph{\emph{Ball Trajectories}}
We generated 700 binary videos of 100 frames each (\(32\times32\)), belonging to seven trajectory classes (horizontal/vertical linear, clockwise/counterclockwise circular, and random). The ball has a circular or elliptical shape, with radii drawn randomly. Trajectories are perturbed by Gaussian noise on the position and uniform jitter, with toroidal wrapping. Additionally, each frame contains independent background noise. Encoding is binary: each active pixel generates a spike in the corresponding frame. Unlike MNIST, this task requires temporal integration and benefits from the reservoir’s fading memory, since class information is distributed over multiple frames.

Dataset sizes are kept limited to make the task more challenging, and this does not affect the analysis, as the focus is on relative performance trends across hyperparameter settings rather than achieving maximum accuracy.

Each spike train is assigned to a randomly chosen internal neuron; each neuron receives at most one input. There is no overlap between input and output neurons, so the input must propagate through the reservoir before reaching the readout.

\subsection{Features and readout layers}
Two types of features are considered: statistical features and trace features.
For the \emph{statistical features}, we use $50$ output neurons for each task. For MNIST, we extract spike count, variance of spike count, time of first spike, and mean spike time from each neuron (4 features per neuron, 200 in total). For Ball Trajectories, we extract mean spike time, time of first and last spike, and the mean and variance of the ISI (5 features per neuron, 250 in total).
For the \emph{trace features}, we choose a number of output neurons (200 for MNIST, 250 for Ball Trajectories) so as to obtain the same total number of features. For each neuron, we consider the value at the end of the simulation of an exponentially filtered spike train with time constant $\tau=60$ for MNIST and $\tau=40$ for Ball Trajectories. The values of $\tau$ were chosen to be consistent with the different durations of the spike trains in the two tasks.

As classifiers, we employ a single-layer perceptron (SLP) with preliminary feature standardization and a Random Forest with 500 trees. 

Performance is estimated via 10-fold stratified cross-validation. For each configuration, we report the mean and standard deviation over the 10 folds for three different metrics $M$: accuracy, F1 with macro averaging (F1--macro), and Matthews Correlation Coefficient (MCC).

\subsection{Experimental Plan}
For each of the 16 experimental settings (2 computational tasks, 2 reset mechanisms, 2 feature types, and 2 readout classifiers), one of the global hyperparameters ($\beta$, $\theta$, $I$) is varied independently over three values, while the remaining ones are kept at their reference values. For every hyperparameter choice, we evaluate the three performance metrics across a range of mean synaptic weights $w$ sampled uniformly in a broad interval centered at the theoretical critical point $w_{\mathrm{crit}}$. We use $w_{\mathrm{crit}}$ only as a reference to define a consistent region of interest across hyperparameter settings.

Throughout the paper, we use the term \emph{trial} to denote one network instance with a full sweep over $w$ with task and reset fixed, evaluating both readouts, both feature types, and all metrics (yielding $2\times2\times3=12$ curves). The connectivity is fixed within a trial, while synaptic weights are resampled at each $w$; a new network instance is generated for each trial. In each hyperparameter sweep, one trial is run for each of the three hyperparameter values; the reference setting is repeated in the $\beta$, $\theta$, and $I$ sweeps (three independent network instances) to estimate trial-to-trial variability.

\subsection{Robustness definition} \label{rob}

\begin{figure}[tb]
    \centering
    \includegraphics[width=1\linewidth]{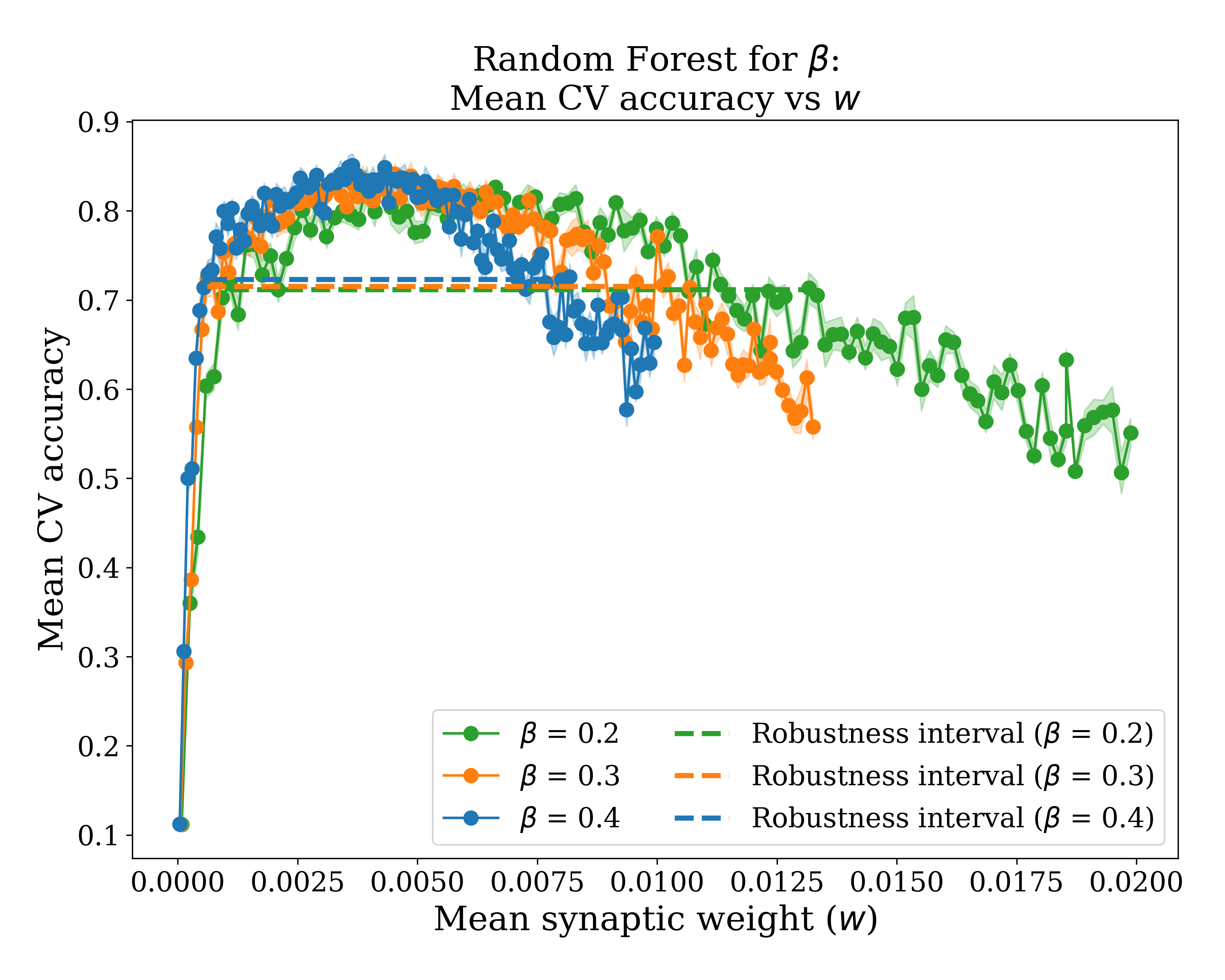}
    \caption{Accuracy--mean synaptic weight curves for MNIST (statistical features, Random Forest, non-fixed reset) and the corresponding robustness intervals for the three values of $\beta$. The shaded area represents the standard deviation.}
    \label{fig:accuracy}
\end{figure}
For each configuration and metric $M$, we obtain the curve $M(w)$ (Fig.~\ref{fig:accuracy}) and define a \emph{relative} threshold
\[
T_{\text{task}}(\gamma)=\gamma\,\max_w M(w),\qquad \gamma\in(0,1).
\]
The \emph{robustness interval} is the range of mean synaptic weights $[w_{\min},w_{\max}]$ for which $M(w)\ge T_{\text{task}}(\gamma)$:
$w_{\min}=\min\{w: M(w)\ge T_{\text{task}}(\gamma)\},$ and 
$w_{\max}=\max\{w: M(w)\ge T_{\text{task}}(\gamma)\},$
with width $\Delta=w_{\max}-w_{\min}$. We sample $w$ on a dense grid around the critical point (from $0.01$ to $2$ times the critical point) defined in Eq.~\eqref{crit}, so that discretization effects are negligible. Larger $\gamma$ increases sensitivity to cross-validation noise near the peak, while smaller $\gamma$ includes low-performance regions. We use $\gamma=0.85$ as the default threshold; trends remain unchanged for $\gamma\in[0.80,0.90]$, thus showing that our findings are not qualitatively affected by this choice.

\subsection{Hyperparameters analysis} \label{analysis}

For each experimental configuration and for each metric $M$, we analyze the curve $M(w)$ as a function of the mean synaptic weight $w$, using the robustness interval and the width $\Delta$ defined in Sec.~\ref{rob} to study how the sensitivity to variations in $w$ changes when $\beta$, $\theta$, and $I$ are varied independently across the different experimental settings. The curves $M(w)$ are obtained as the average over the 10 cross-validation folds and are used in their discrete form, without any smoothing or fitting; the robustness intervals are therefore computed directly from the empirical performance values for each sampled $w$. 

As a topology control, we repeat the same procedure under directed Erd\H{o}s--R\'enyi connectivity for a representative subset of settings (MNIST; both feature types and readouts; variations in $\beta$ and $\theta$).

\medskip
\noindent\textbf{Theoretical firing rate.}
In addition to the empirical curves, we also use a theoretical prediction obtained from a mean-field approach, based on the expression for the mean ISI derived in \cite{freddi2025meanfield}:
\[
\text{ISI}_{\text{mean}} =
\frac{1}{2I}\Bigl(\theta - w \beta N
+ \sqrt{
(\theta - w \beta N)^2
+ 4 I \beta N T_{\mathrm{ref}} w
}\,\Bigr),
\]
from which we obtain the theoretical firing rate
\begin{equation} \label{fr}
\nu_{\mathrm{th}}(w) = \frac{1}{\text{ISI}_{\text{mean}}}.
\end{equation}
We analyze how $\nu_{\mathrm{th}}(w)$ varies with the hyperparameters ($\beta$, $\theta$, $I$) to qualitatively anticipate the expected changes in the width of the robustness interval, and we compare these predictions with the data obtained from simulations.

\medskip
\noindent\textbf{Critical mean synaptic weight.}
We also use the mean-field estimate of the critical mean synaptic weight proposed in \cite{freddi2025meanfield}:
\begin{equation} \label{crit}
 w_{\mathrm{crit}} =
\frac{\theta - 2 I T_{\mathrm{ref}}}{\beta N}.   
\end{equation}
For each experimental configuration, we evaluate whether the predicted critical value falls within the robustness interval, as an indicator of the accuracy of the theoretical approximation. We also assess whether this value lies to the left of the midpoint of the robustness interval, in order to quantify how often the interval is biased toward sub/supercritical regimes.

\medskip
\noindent\textbf{Equivalent Hyperparameters.} \label{pareq}
To explore transformations that preserve the critical point, we start from the reference values $(\beta_{\mathrm{ref}},\theta_{\mathrm{ref}})$ defined in Sec.~\ref{first} and define an equivalent firing threshold $\theta_{\mathrm{eq}}$ by imposing
\[
w_{\mathrm{crit}}(\beta_{\mathrm{alt}}, \theta_{\mathrm{ref}})
=
w_{\mathrm{crit}}(\beta_{\mathrm{ref}}, \theta_{\mathrm{eq}}), 
\]
from which we obtain
\begin{equation} \label{equiv}
  \theta_{\mathrm{eq}} =
\left( \theta_{\mathrm{ref}} - 2 I_{\mathrm{ref}} T_{\mathrm{ref}} \right)
\frac{\beta_{\mathrm{ref}}}{\beta_{\mathrm{alt}}}
+ 2 I_{\mathrm{ref}} T_{\mathrm{ref}}.  
\end{equation}

For each pair of alternative configurations $(\beta_{\mathrm{alt}}, \theta_{\mathrm{ref}})$ and $(\beta_{\mathrm{ref}}, \theta_{\mathrm{eq}})$, we compare both the theoretical firing rate curves $\nu_{\mathrm{th}}(w)$ and the empirical curves $M(w)$, in order to assess the extent to which the transformation preserves the dependence of performance on the mean synaptic weight.

\medskip
\noindent\textbf{Limit of the method for the input $I$.}
When applying the same procedure to the input $I$, the resulting values of $I_{\mathrm{alt}}$ become incompatible with the model dynamics, as they exceed the firing threshold. In principle, this issue could be mitigated by selecting values of $\beta_{\mathrm{alt}}$ very close to the reference $\beta_{\mathrm{ref}}$, but this would trivialize the hyperparameter variation. For this reason, we do not consider equivalent transformations based on $I$.

\section{Results}
\subsection{Firing rate theoretical behavior}

\begin{figure}[tb]
    \centering

    \begin{minipage}[t]{0.49\linewidth}
        \centering
        \includegraphics[width=\linewidth]{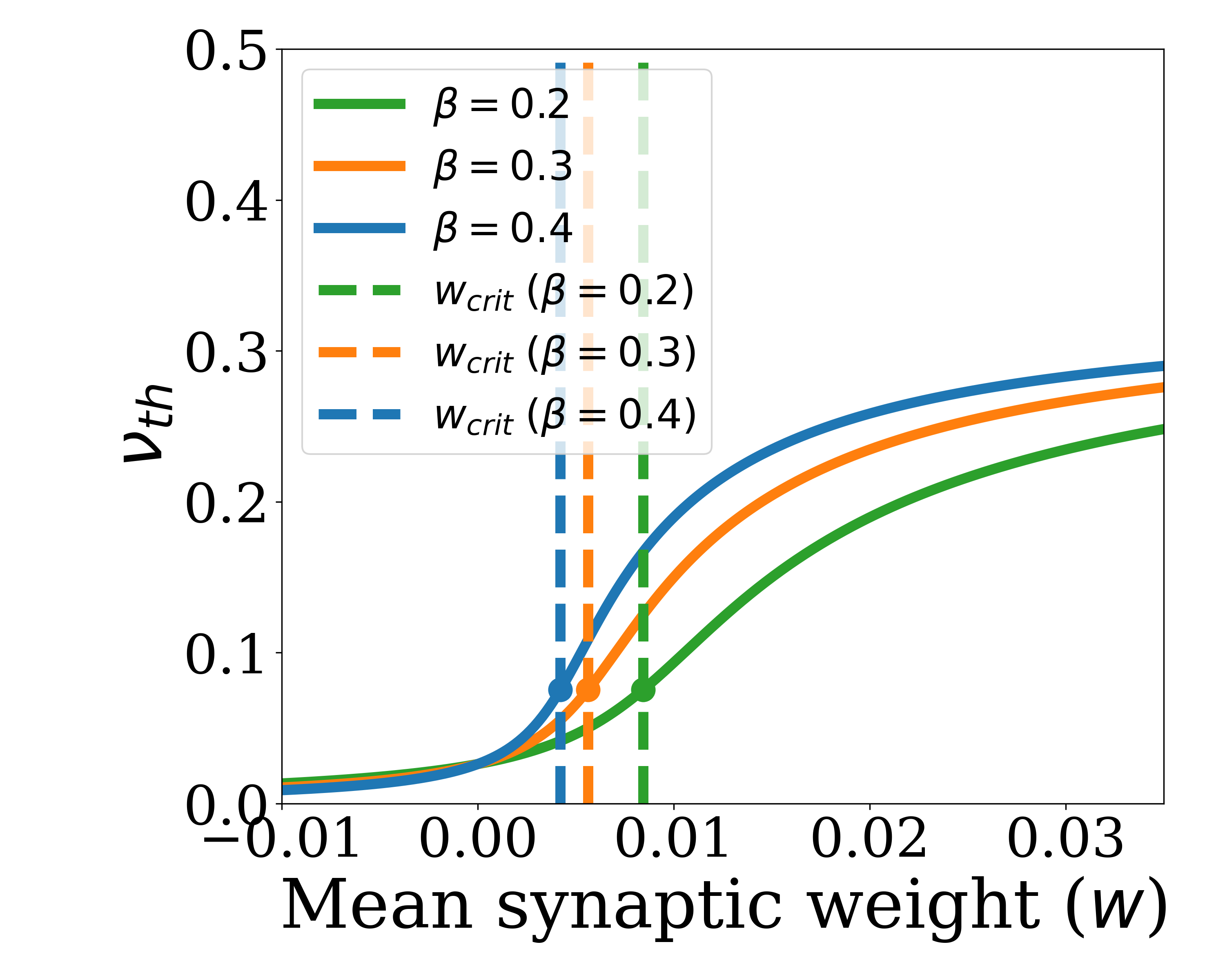}
        {\footnotesize (a) Theoretical curve $\nu_{\mathrm{th}}(w)$ for the MNIST task as a function of $\beta$.}
    \end{minipage}
    \hfill
    \begin{minipage}[t]{0.49\linewidth}
        \centering
        \includegraphics[width=\linewidth]{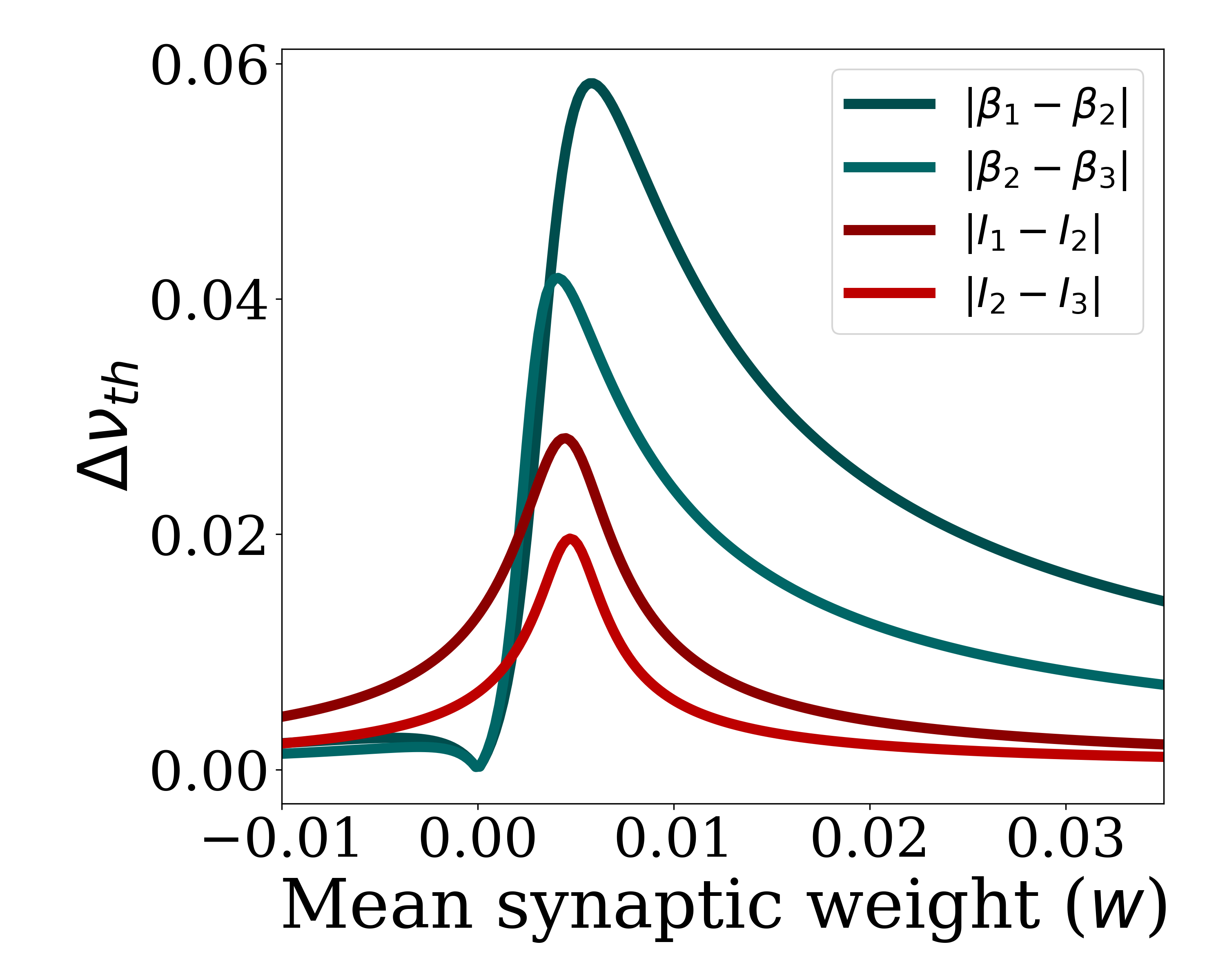}
        {\footnotesize (b) Absolute value of the difference between firing-rate curves for the MNIST task: 
variation in $\beta$ (blue) vs.\ variation in $I$ (red).
}
    \end{minipage}

    \caption{Effect of $\beta$ and $I$ on the theoretical firing-rate curve.}
    \label{fig:fr_beta_I}
\end{figure}

Fig.~\ref{fig:fr_beta_I}(a) shows the theoretical firing rate curve $\nu_{\mathrm{th}}(w)$ as a function of the mean synaptic weight $w$, obtained from the mean-field expression in Eq.~\eqref{fr}, for three different values of $\beta$ (with all other hyperparameters fixed at the reference configuration). Increasing $\beta$ steepens the transition between the quiescent and saturated regimes, thereby reducing the range of weights for which the firing rate takes on intermediate and potentially useful values for the reservoir. Since the experimental robustness interval must lie within this range, a steeper theoretical curve implies a narrower robustness interval as $\beta$ increases. Conversely, increasing the firing threshold $\theta$ is expected to have the opposite effect, flattening the curve and extending the range of intermediate firing rates. 
The changes introduced by the input \(I\) are qualitatively similar to those of $\theta$, but more limited: the form of \(\nu_{\mathrm{th}}(w)\) varies only slightly and almost exclusively near the critical point \(w_{\mathrm{crit}}\). To quantify these differences, we consider the pointwise variation of each curve relative to the reference configuration (Fig.~\ref{fig:fr_beta_I}(b)). The differences are more evident for \(\beta\) and \(\theta\), while they remain smaller for \(I\), confirming that the theoretical sensitivity of the dynamics to the input \(I\) is intrinsically weaker; moreover, the range of values it can take is limited, as discussed in Sec.~\ref{first}.

\subsection{Robustness interval width: experimental findings}
The empirical analysis covers all \(16\) experimental configurations and, for each of them,
the \(3 \times 3 = 9\) robustness intervals obtained from the three performance metrics and the
three values of the varied hyperparameter, for a total of \(144\) intervals examined for each hyperparameter.
To correctly interpret the results, it is useful to quantify the intrinsic variability of the
curves \(M(w)\): for each value of \(w\), we compute the coefficient of variation of \(M(w)\) 
and take its average across all points and all configurations. According to this measure, the Ball Trajectories task is on average about 6–7 times noisier than MNIST, which makes the weaker effects predicted by the theory harder to detect experimentally and suggests that local deviations from the theoretical expectations are mainly due to task-induced noise rather than to systematic discrepancies in the model. However, within this broader perspective, data shows a clear overall consistency with the theoretical predictions based on the firing-rate curve $\nu_{\mathrm{th}}(w)$.

\begin{figure}[tb]
    \centering
    \includegraphics[width=0.8\linewidth]{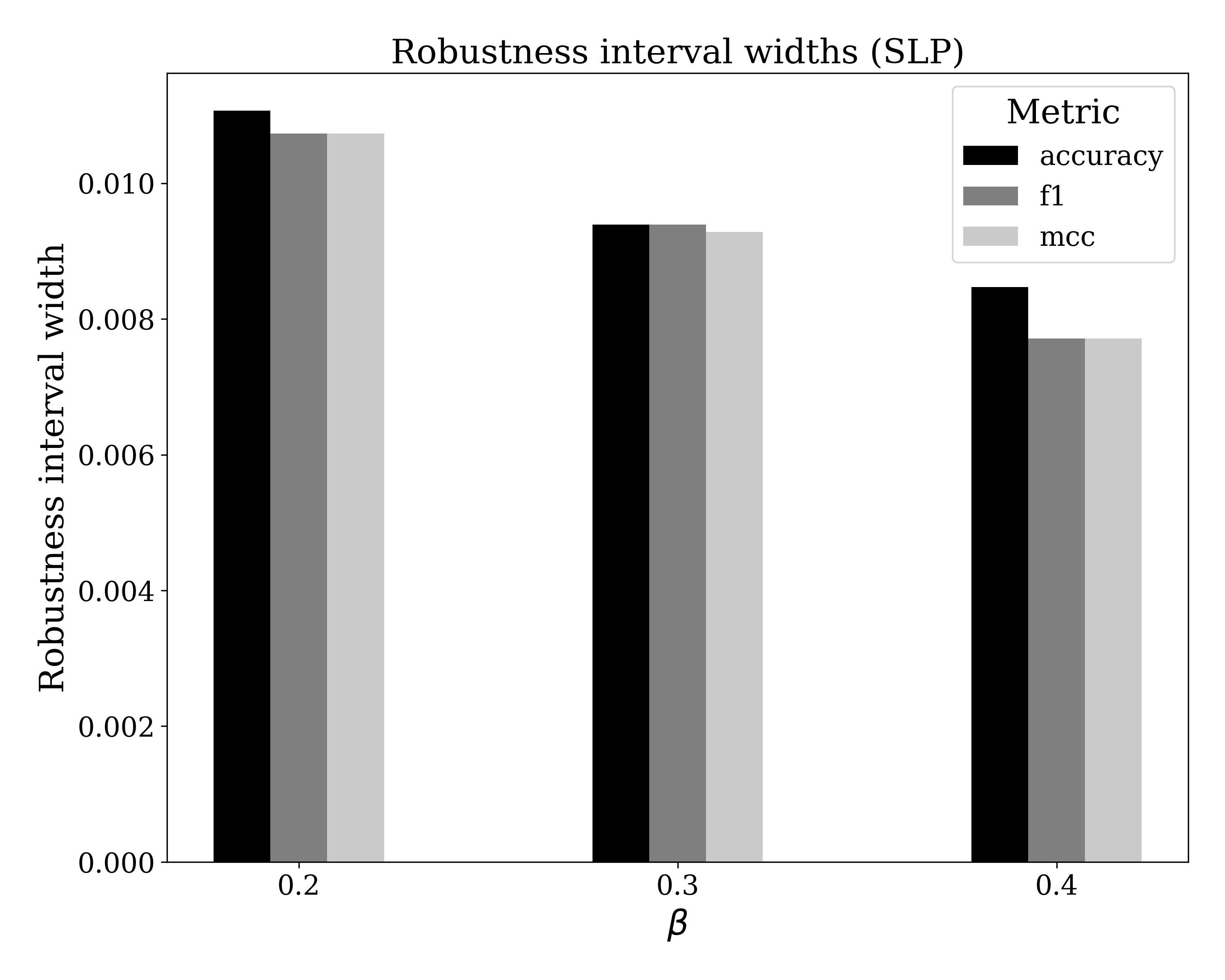}
    \caption{Amplitude of robustness interval $\Delta$ as $\beta$ varies for the three metrics (accuracy, F1-macro, MCC). Data related to the MNIST task with SLP readout, statistical features and fixed reset.}
    \label{fig:delta_beta_slp}
\end{figure}

\paragraph*{Effect of $\beta$}
In \(97.9\%\) of the intervals examined, \(\Delta\) decreases as \(\beta\) increases (i.e. it increases as sparsity increases), with a uniform trend across all settings and in agreement with the theoretical predictions. Fig.~\ref{fig:delta_beta_slp} shows a representative example, in which the monotonic decrease of \(\Delta\) with \(\beta\) is particularly clear for all three metrics. The trend observed in the figure is qualitatively identical to what is found in all other experimental settings.

\paragraph*{Effect of $\theta$}
The influence of the firing threshold is equally clear: in \(94.8\%\) of the cases, the width \(\Delta\) increases as \(\theta\) grows, in agreement with the predicted flattening of the theoretical firing–rate curve. The consistency between theory and simulations is comparable to that observed for \(\beta\).

\paragraph*{Effect of $I$}
Variations in \(I\) produce weaker effects, as anticipated by the theoretical
analysis. Across all
experimental configurations, the dependence of \(\Delta\) on \(I\) matches the
theoretical expectation in \(84.4\%\) of the examined cases.

Table~\ref{tab:param_relative_changes} summarizes the relative average variation of the robustness-interval amplitude. The effects associated with $\beta$ and $\theta$ are markedly more pronounced than those produced by $I$, in agreement with the mean-field predictions.

\begin{table}[tb]
    \centering
    \caption{The reported quantities are the averages of the relative differences of the robustness interval width (where e.g. “mid-min” for $\beta$ refers to the second largest value of $\beta$ and the minimum value) for accuracy, F1-macro, and MCC as the reservoir hyperparameters ($\beta,\theta, I$) vary.}
    \label{tab:param_relative_changes}
    \resizebox{\linewidth}{!}{%
    \setlength{\tabcolsep}{3pt}
    \begin{tabular}{lcccccc}
        \hline
        Hyperpar. &
        acc\_mid\_min &
        acc\_max\_mid &
        f1\_mid\_min &
        f1\_max\_mid &
        mcc\_mid\_min &
        mcc\_max\_mid \\
        \hline
        $\beta$  & -0.2647 & -0.1876 & -0.2509 & -0.1910 & -0.2387 & -0.1679 \\
        $\theta$ &  0.2076 &  0.1829 &  0.2093 &  0.1846 &  0.1683 &  0.2641 \\
        $I$      &  0.1629 &  0.1067 &  0.1403 &  0.1194 &  0.1827 &  0.1529 \\
        \hline
    \end{tabular}%
    }
\end{table}

\subsection{Equivalent Hyperparameters $(\beta,\theta)$}

\begin{figure}[tb]
    \centering
    \includegraphics[width=0.9\linewidth]{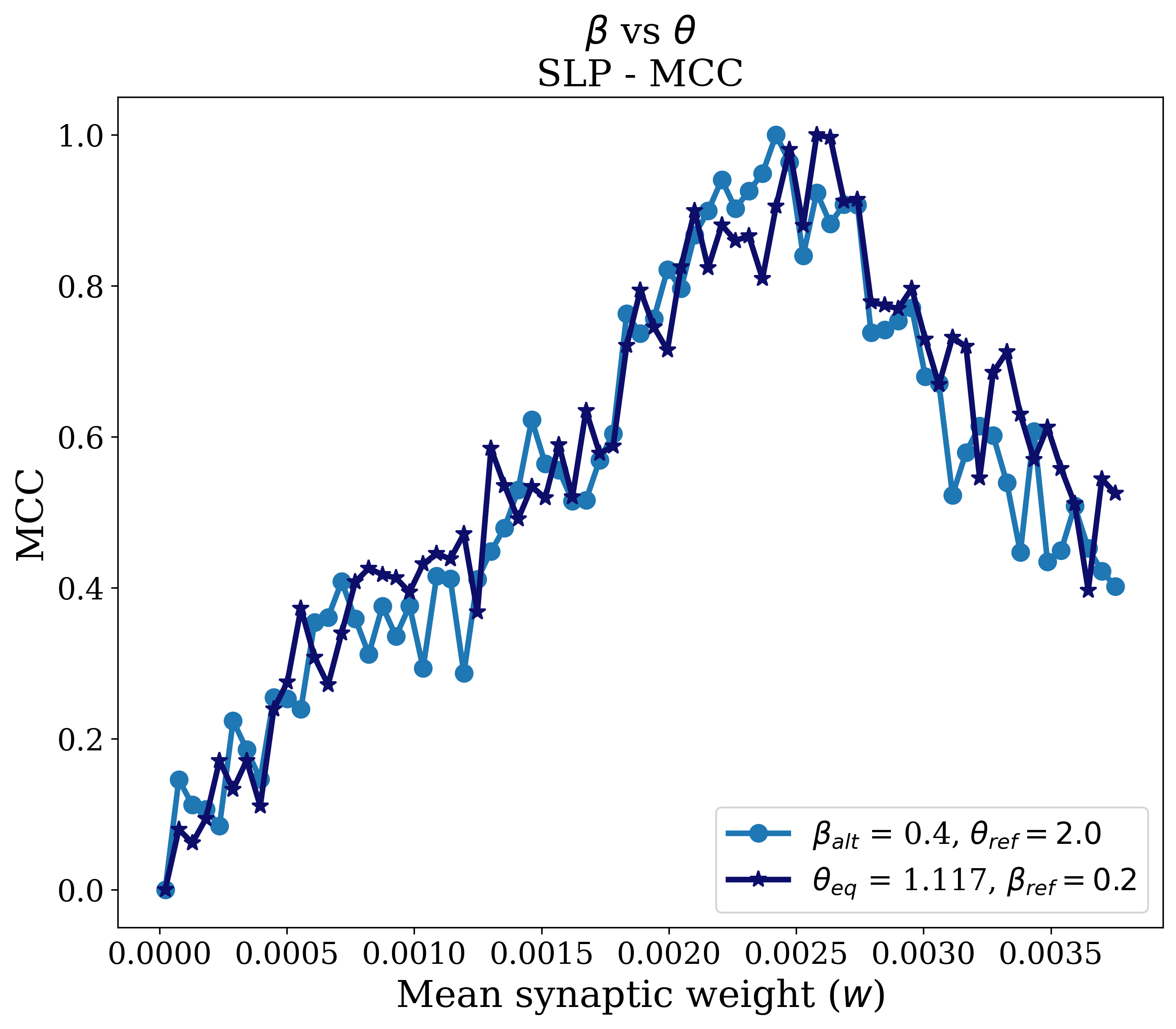}
    \caption{Comparison between the accuracy curves $M(w)$ for the Ball Trajectories task (statistical features, Single Layer Perceptron readout, non-fixed reset). The configurations $(\beta_{\mathrm{alt}}, \theta_{\mathrm{ref}})$ are compared with the corresponding equivalent configurations $(\beta_{\mathrm{ref}}, \theta_{\mathrm{eq}})$ that preserve the theoretical critical point. The resulting curves are very similar, with deviations falling within the range of inter-trial variability.}
    \label{fig:equiv}
\end{figure}

We consider alternative values $\beta_{\mathrm{alt}} \neq \beta_{\mathrm{ref}}$ and use the reference hyperparameter values introduced in Sec.~\ref{first}. For each of these, we use the equivalent firing threshold $\theta_{\mathrm{eq}}$ defined in Eq.~\eqref{equiv} to preserve the theoretical value of the critical point.

\paragraph*{Theoretical similarity}
For the pairs of configurations
\(
(\beta_{\mathrm{alt}}, \theta_{\mathrm{ref}})\) and \(
(\beta_{\mathrm{ref}}, \theta_{\mathrm{eq}})
\)
the corresponding theoretical firing rate curves $\nu_{\mathrm{th}}(w)$ are nearly
overlapping: besides the critical point, the overall shape of the curve remains
practically unchanged.  We quantify this similarity using the normalized distance
\begin{equation} \label{norm}
  d =
\frac{\int \bigl|\nu_{\mathrm{th}}^1(w) - \nu_{\mathrm{th}}^2(w)\bigr|\,dw}
     {\tfrac{1}{2}\int \bigl(\nu_{\mathrm{th}}^1(w) + \nu_{\mathrm{th}}^2(w)\bigr)\,dw},  
\end{equation}
where the integral is calculated over the considered interval of weights (see Sec.~\ref{rob}) and $\nu_{\mathrm{th}}^1$ and $\nu_{\mathrm{th}}^2$ are the firing rates corresponding to the two different configurations. The values for $d$ among the different settings range from $0.0011$ to $0.004$.

\paragraph*{Empirical similarity}
The same equivalence is reflected in the empirical curves $M(w)$. The curves in
Fig.~\ref{fig:equiv} show modest differences, comparable to the variability between trials. We define   $d_{\mathrm{norm}}$ as in Eq.\eqref{norm}, where we use $M^{\mathrm{norm}}_1$ and $M^{\mathrm{norm}}_2$ instead of $\nu_{\mathrm{th}}^1$ and $\nu_{\mathrm{th}}^2$ and where $M^{\mathrm{norm}}(w)$ denotes the curve $M(w)$ after linear normalization of
the y-axis, so that the maximum value is 1 (since we are interested in the shape of
the curve rather than the maximum accuracy). We find that $d_{\mathrm{norm}}$ computed across pairs of trials with identical hyperparameters (independent network instances) and across pairs of trials with equivalent hyperparameters has nearly identical mean values ($d_{\mathrm{norm}}\approx 0.055$ in both cases), and ANOVA does not detect significant differences ($p>0.95$). This indicates that the normalized curves have essentially the same shape.

\subsection{Topology control (Erd\H{o}s--R\'enyi)}
To assess the dependence of our findings on the connectivity structure, we repeated the experiments under directed Erd\H{o}s--R\'enyi random graphs for the subset described in Sec.~\ref{analysis}. The monotonic trends of the robustness-interval width $\Delta$ with respect to $\beta$ (decreasing) and $\theta$ (increasing) were verified in $\sim$100\% of the examined cases. Moreover, the $(\beta,\theta)$ equivalence produced nearly overlapping normalized performance curves, with an average normalized distance (Eq.~\eqref{norm}) of order $\sim 3\times10^{-2}$, consistent with the small-world results.

\subsection{Consistency of the criticality reference}
For each experimental configuration, we evaluate, as described in Eq.~\eqref{crit}, whether the theoretical value \(w_{\mathrm{crit}}\) falls within
the empirical robustness interval and report the corresponding agreement rates.

Across all experimental settings, the theoretical value lies within the robustness interval:
\begin{itemize}
    \item in 98\% of cases for accuracy,
    \item in 95.1\% of cases for F1--macro,
    \item in 90.2\% of cases for MCC.
\end{itemize}
These percentages suggest that the mean-field approximation accurately identifies the region of synaptic weights in which the reservoir maintains high and stable performance.

Stratifying by dataset, we observe perfect agreement for MNIST (100\% for all metrics), whereas Ball Trajectories yields lower agreement rates, consistent with its higher input and temporal variability.

In addition, we observe that the center of the robustness interval lies consistently below \( w_{\mathrm{crit}} \) (the interval was always found to start no earlier than $w_{\mathrm{crit}}/50$ and end no later than $2\,w_{\mathrm{crit}}$), so that high-performance weights are more often found in a mildly subcritical regime. This empirical bias is in line with the asymmetric shape of the theoretical curve \(\nu_{\mathrm{th}}(w)\), which varies more gradually on the subcritical side and more steeply on the supercritical side.

\section{Discussion and Conclusions}
Given the inherent uncertainty in practical reservoir computing applications, this work advocates for a shift from a point-wise view of criticality toward an operational range perspective. Focusing on LIF-based spiking architectures, we provide a characterization of the robustness-interval (i.e. the safety margin available for reservoir tuning) morphology. 
A first central finding is that mean-field firing-rate theory accurately captures the empirical trends: the robustness-interval width varies predictably with connectivity ($\beta$), firing threshold ($\theta$) and, less strongly, with the input values. Specifically, from a practical design perspective, sparser networks or higher firing thresholds yield wider robustness intervals, thereby reducing sensitivity to fine tuning of the mean synaptic weight $w$. 

Furthermore, the empirical validation of the theoretical critical point $w_{\mathrm{crit}}$ (Eq.~\eqref{crit}) strengthens confidence in its use as an operational reference. Since the empirical optimum could in principle lie anywhere within the explored interval, the observation that $w_{\mathrm{crit}}$ falls within the robustness interval (the interval always fell within the range $[w_{crit} / 50, 2 w_{crit}]$) provides \emph{a posteriori} evidence that the mean-field estimate is a useful starting coordinate for search and fine-tuning, rather than a target operating point. We also observe that the center of the robustness interval consistently lies below $w_{\mathrm{crit}}$, so that the most reliable operating region is typically mildly subcritical. This systematic shift is consistent with the asymmetry of the theoretical firing--rate curve $\nu_{\mathrm{th}}(w)$, which favors a broader range of usable weights on the subcritical side.

Finally, we identify equivalent parameter pairs $(\beta,\theta)$ that preserve the theoretical critical point and yield nearly overlapping performance curves 
$M(w)$, enabling the same operating regimes with different global-parameter combinations and thus greater design flexibility when one parameter is constrained. This may reflect a functional degeneracy, distinct configurations producing similar computational behavior, akin to degeneracy in biological neural circuits \cite{marder2006variability}.

This work has some limitations by design. The dynamics are based on LIF neurons with fixed synaptic weights, which enables controlled and directly comparable analyses of robustness but excludes adaptive mechanisms such as synaptic plasticity or homeostatic regulation. Concerning network structure, the analysis focuses on small-world connectivity, with a topology control on directed Erd\H{o}s--R\'enyi random graphs confirming the main trends for the considered subset; however, a broader exploration of more diverse connectivity families could further strengthen the generality of the conclusions.

Future work may include more biologically realistic neuronal models, modular or hierarchical architectures, and synaptic plasticity. It will also extend the analysis to more diverse datasets (e.g., sMNIST, Lorenz96), explore sparser connectivity regimes (e.g., $\beta < 0.1$), and further refine the theoretical characterization of the robustness interval.

In summary, the results concerning the robustness-interval width, equivalent $(\beta,\theta)$ pairs, and an existing estimate of the critical point, provide practical guidelines to set global reservoir parameters and select robust operating ranges with limited tuning effort.

\section*{Acknowledgements}
AB is supported by the European Research Council (ERC Synergy) under the European Union’s Horizon 2020 research and innovation programme (ConnectToBrain; Grant Agreement No. 810377). The content of this article reflects only the author’s view and the ERC Executive Agency is not responsible for the content.


\begin{thebibliography}{1}

\bibitem{zhang2023survey}
H.~Zhang and D.~V.~Vargas,
``A Survey on Reservoir Computing and Its Interdisciplinary Applications Beyond Traditional Machine Learning,''
\emph{IEEE Access}, vol.~11, pp.~81033--81078, 2023.

\bibitem{alvarez2025optimizing}
O.~I.~Alvarez-Canchila, A.~Espinal, A.~Pati\~no-Saucedo, and H.~Rostro-Gonzalez,
``Optimizing Reservoir Separability in Liquid State Machines for Spatio-Temporal Classification in Neuromorphic Hardware,''
\emph{Journal of Low Power Electronics and Applications}, vol.~15, no.~1, p.~4, 2025.

\bibitem{maass2002liquid}
W.~Maass, T.~Natschl\"ager, and H.~Markram,
``Real-time computing without stable states: A new framework for neural computation based on perturbations,''
\emph{Neural Computation}, vol.~14, no.~11, pp.~2531--2560, 2002.

\bibitem{alvarez2025separability}
O.~I.~Alvarez-Canchila, A.~Espinal, A.~Pati\~no-Saucedo, and H.~Rostro-Gonzalez,
``Enhancing Liquid State Machine Classification Through Reservoir Separability Optimization Using Swarm Intelligence and Multitask Learning,''
\emph{Neural Computing and Applications}, 2025.

\bibitem{zhang2015digitallsm}
Y.~Zhang, P.~Li, Y.~Jin, and Y.~Choe,
``A digital liquid state machine with biologically inspired learning and its application to speech recognition,''
\emph{IEEE Transactions on Neural Networks and Learning Systems}, vol.~26, no.~11, pp.~2635--2649, 2015.


\bibitem{ivanov2021edge}
V.~A.~Ivanov and K.~P.~Michmizos,
``Increasing Liquid State Machine Performance with Edge-of-Chaos Dynamics Organized by Astrocyte-Modulated Plasticity,''
in \emph{Proc. NeurIPS}, 2021.

\bibitem{woo2024characterization}
J.~Woo, S.~H.~Kim, H.~Kim, and K.~Han,
``Characterization of the neuronal and network dynamics of liquid state machines,''
\emph{Physica A: Statistical Mechanics and its Applications}, vol.~633, p.~129334, 2024.

\bibitem{bertschinger2004edge}
N.~Bertschinger and T.~Natschl\"ager,
``Real-time computation at the edge of chaos in recurrent neural networks,''
in \emph{Advances in Neural Information Processing Systems (NIPS)}, vol.~17,
pp.~1451--1458, 2004.

\bibitem{brochini2016critical}
L.~Brochini, A.~A.~Costa, M.~Abadi, A.~Roque, J.~Stolfi, and O.~Kinouchi,
``Phase transitions and self-organized criticality in networks of stochastic spiking neurons,''
\emph{Scientific Reports}, vol.~6, p.~35831, 2016.

\bibitem{levina2007soc}
A.~Levina, J.~M.~Herrmann, and T.~Geisel,
``Dynamical synapses causing self-organized criticality in neural networks,''
\emph{Nature Physics}, vol.~3, no.~12, pp.~857--860, 2007.

\bibitem{schrauwen2008robustness}
B.~Schrauwen, M.~Wardermann, D.~Verstraeten, J.~Steil, and D.~Stroobandt,
``Improving reservoirs using intrinsic plasticity,''
\emph{Neurocomputing}, vol.~71, no.~7--9, pp.~1159--1171, 2008.

\bibitem{hazan2010lsmrobust}
H. Hazan and L. Manevitz,
``The liquid state machine is not robust to problems in its components but topological constraints can restore robustness,''
in \emph{Proc. Int. Conf. on Fuzzy Computation and Int. Conf. on Neural Computation (ICFC)}, 2010, pp. 258--264.

\bibitem{kinouchi2006optimal}
O.~Kinouchi and M.~Copelli,
``Optimal dynamical range of excitable networks at criticality,''
\emph{Nature Physics}, vol.~2, no.~5, pp.~348--351, 2006.

\bibitem{freddi2025meanfield}
R.~Freddi, F.~Cicala, L.~Marzetti, and A.~Basti,
``A mean-field approach to criticality in spiking neural networks for reservoir computing,''
\emph{Scientific Reports}, vol.~15, no.~1, p.~34709, 2025.

\bibitem{watts1998collective}
D.~J.~Watts and S.~H.~Strogatz,
``Collective dynamics of ``small-world'' networks,''
\emph{Nature}, vol.~393, no.~6684, pp.~440--442, 1998.

\bibitem{marder2006variability}
E.~Marder and J.-M.~Goaillard,
``Variability, compensation and homeostasis in neuron and network function,''
\emph{Nature Reviews Neuroscience}, vol.~7, no.~7, pp.~563--574, 2006.


\end{thebibliography}
\end{document}